\pdfoutput=1

\documentclass[11pt]{article}

\usepackage{emnlp2021}

\usepackage{times}
\usepackage{latexsym}

\usepackage{sec/package}
\usepackage{url}
\usepackage{hyperref}

\newcommand\myparagraph[1]{
\vskip 0.05in 
\noindent{\bf {#1}}}

\usepackage[T1]{fontenc}

\usepackage[utf8]{inputenc}

\usepackage{microtype}

\usepackage{notation}
\usepackage{cleveref}
\crefname{figure}{Fig.}{Figs.}
\crefname{table}{Tab.}{Tabs.}
\crefname{section}{\S}{\S\S}
\crefname{subsection}{\S}{\S\S}
\crefname{subsubsection}{\S}{\S\S}

\title{
Causal Direction of Data Collection Matters:\\
Implications of Causal and Anticausal Learning for NLP

}

\author{
Zhijing Jin\textsuperscript{1,2}\thanks{ { } Equal contribution.} \,,
Julius von K\"ugelgen\textsuperscript{1,3}\footnotemark[1] \,, Jingwei Ni\textsuperscript{4},
Tejas Vaidhya\textsuperscript{5}, Ayush Kaushal\textsuperscript{5}, \\
{\bf Mrinmaya Sachan\textsuperscript{2} \and Bernhard Sch\"olkopf\textsuperscript{1,2}}
\\
\textsuperscript{1}Max Planck Institute for Intelligent Systems, Tübingen, Germany,
\\ 
\textsuperscript{2}ETH Zürich,
\textsuperscript{3}University of Cambridge, \textsuperscript{4}University College London, \textsuperscript{5}IIT Kharagpur\\
\texttt{\{zjin,jvk,bs\}@tue.mpg.de, ucabjni@ucl.ac.uk}, \\
\texttt{\{iamtejasvaidhya,ayushkaushal\}@iitkgp.ac.in, msachan@ethz.ch }
}

\begin{document}
\maketitle

\begin{abstract}
The principle of independent causal mechanisms (ICM) states that generative processes of real world data consist of independent modules which do not influence or inform each other. While this idea has led to fruitful developments in the field of causal inference, it is not widely-known in the NLP community. In this work, we argue that the causal direction of the data collection process bears nontrivial implications that can explain a number of published NLP findings, such as differences in semi-supervised learning (SSL) and domain adaptation (DA) performance across different settings. We categorize common NLP tasks according to their causal direction and empirically assay the validity of the ICM principle for text data using minimum description length. We conduct an extensive meta-analysis of over 100 published SSL and 30 DA studies, and find that the results are consistent with our expectations based on causal insights. This work presents the first attempt to analyze the ICM principle in NLP, and provides constructive suggestions for future modeling choices.\footnote{Code available at \href{https://github.com/zhijing-jin/icm4nlp}{https://github.com/zhijing-jin/icm4nlp}.}
\end{abstract}

\section{Introduction}\label{sec:intro}
NLP practitioners typically do not pay great attention to the causal direction of the data collection process.
As a motivating example, consider the case of collecting a dataset to train a machine translation (MT) model to translate from
English (En) to Spanish (Es): 
it is common practice to
mix all available En-Es sentence pairs together and train 
the model on the entire pooled data set~\cite{bahdanau2014neural,cho2014learning}. 
However, such mixed corpora actually consist of two distinct types of data: (i)
sentences that originated in English and have been translated (by human translators) into Spanish (En$\rightarrow$Es); and (ii) 
sentences that originated in Spanish and have subsequently been translated 
into English (Es$\rightarrow$En).\footnote{There is, in principle, a third option: both could be translations from a third language, but this occurs less frequently.}

Intuitively, these two subsets are qualitatively different, and 
an increasing number of observations by the NLP community indeed suggests that 
they exhibit
different properties~\cite{freitag-etal-2019-ape,edunov-etal-2020-evaluation,riley-etal-2020-translationese,shen-etal-2021-source}.
In the case of MT, for example, researchers find that training models 
on each of these two types of data separately leads to
different test performance, as well as different performance improvement by semi-supervised learning (SSL)~\cite{bogoychev2019domain,graham-etal-2020-statistical,edunov-etal-2020-evaluation}.
Motivated by this observation that the data collection process seems to matter for model performance, in this work, we provide an explanation of this
phenomenon
from the perspective of causality~\cite{pearl2009causality,peters2017elements}.

\begin{figure}[t]
    \centering
    \includegraphics[width=\columnwidth]{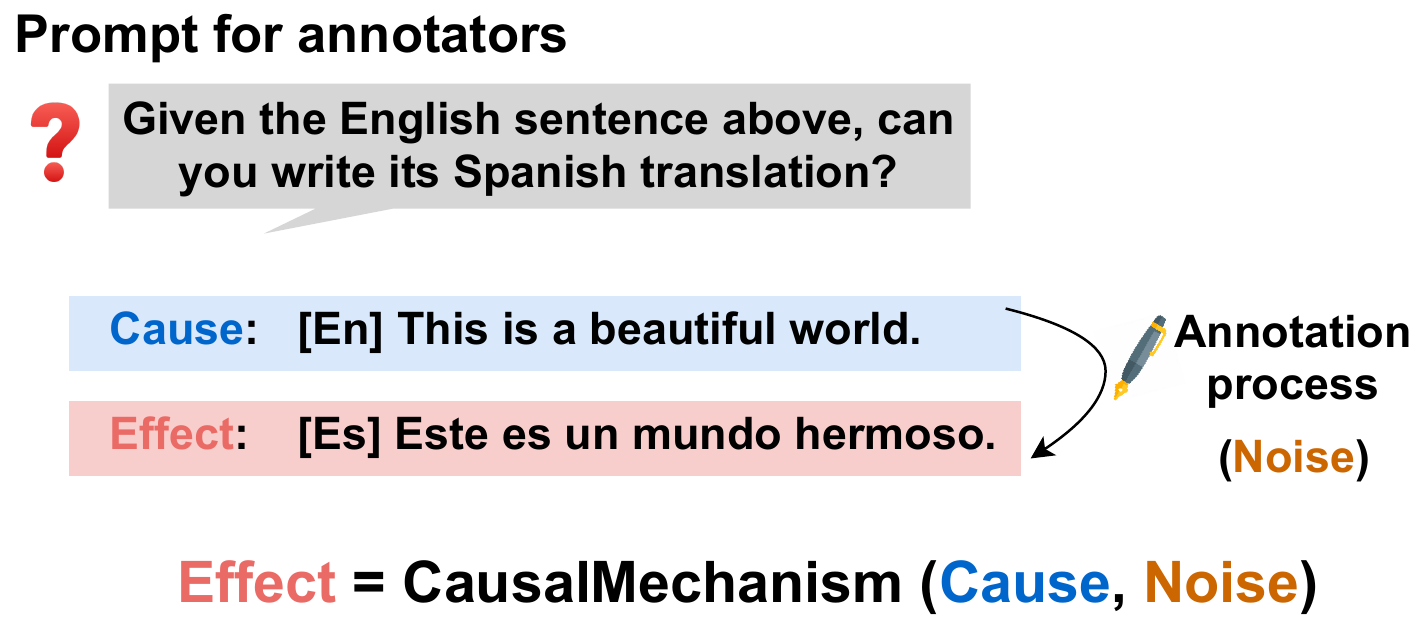}
    \vspace{-1em}
    \caption{Annotation process for NLP data: the random variable that exists first is typically the cause (e.g., a given prompt), and the one generated afterwards is typically the effect (e.g., the annotated answer).}
    \label{fig:prompt}
    \vspace{-0.5em}
\end{figure}
First, we introduce the notion of the \textit{causal direction} for a given NLP task, see~\cref{fig:prompt} for an example.
Throughout, we denote the
input of a learning task by~$X$ and the output which is to be predicted by~$Y$.
If, during the data collection process,  $X$ is generated first, and then $Y$ is collected based on
$X$ (e.g., through annotation), we say that $X$ causes $Y$, and denote this by $X\rightarrow Y$.
If, on the other hand,
 $Y$ is generated first, and then $X$ is collected based on $Y$, we say that $Y$ causes $X$ ($Y\rightarrow X$).\footnote{This corresponds to an \textit{interventional} notion of causation: if one were to manipulate the cause,  the annotation process would lead to a potentially different effect. A manipulation of the effect, in contrast, would not change the cause.}

\begin{figure}[t]
    \vspace{-0.5em}
    \centering
    \includegraphics[width=\columnwidth]{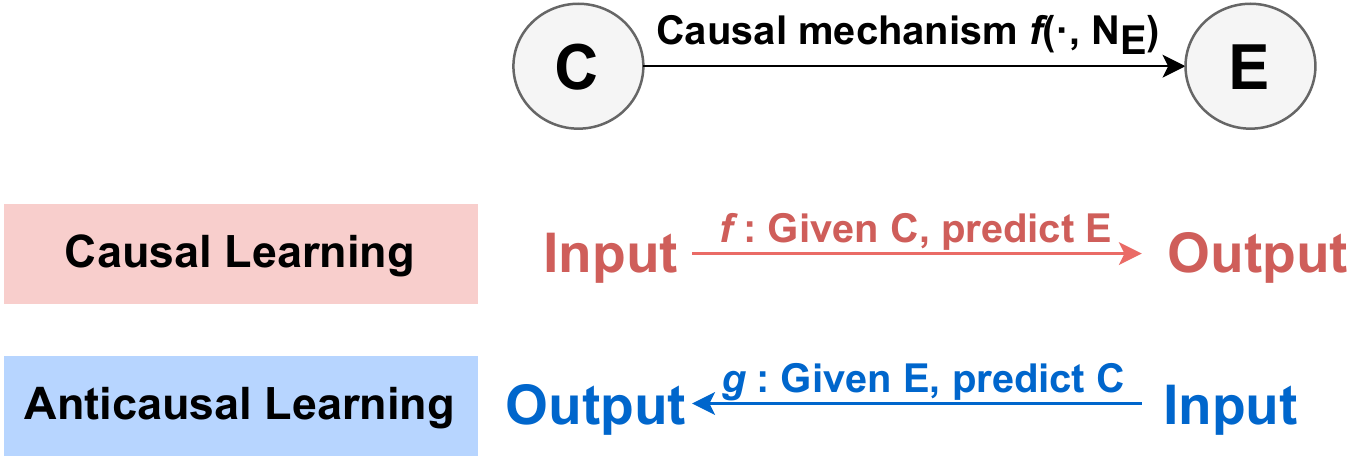}
    \vspace{-1em}
    \caption{\textit{(Top)} A causal graph $C \rightarrow E$, where $C$ is the cause and $E$ is the effect. The function $f(\cdot, N_E)$ denotes the causal process, or mechanism, $P_{E|C}$ by which the effect $E$ is generated from $C$ and  unobserved noise $N_E$. 
    \textit{(Bottom)} Based on whether the direction of prediction aligns with the direction of causation or not, we distinguish two types of tasks: (i)
    \myred{causal learning}, i.e., predicting the effect from the cause;
    and (ii)
    \myblue{anticausal learning}, i.e., predicting the cause from the effect.
    }
    \label{fig:causal_graph}
    \vspace{-0.5em}
\end{figure}

Based on whether the direction of prediction
aligns with the causal direction of the data collection process or not, 
 \citet{schoelkopf2012causal} categorize these types of tasks as \textit{causal learning} ($X\rightarrow Y$), or \textit{anticausal learning} ($Y\rightarrow X$), respectively; see~\cref{fig:causal_graph} for an illustration.
 In the context of our motivating MT example
 this means that,
 if the  goal is to translate from English ($X=\text{En}$) 
 into Spanish ($Y=\text{Es}$),  training \textit{only} on subset (i) of the data
 consisting of En$\rightarrow$Es pairs 
 corresponds to \textit{causal learning} ($X\rightarrow Y$), whereas training \textit{only} on subset (ii) consisting of Es$\rightarrow$En pairs is categorised as \textit{anticausal learning} ($Y\rightarrow X$).

Based on the principle of independent causal mechanisms (ICM)~\cite{janzing2010causal,peters2017elements}, it has been hypothesized that the causal direction of data collection (i.e., whether a given NLP learning task can be classified as causal or anticausal) has implications for the effectiveness of commonly used techniques such as SSL and domain adaptation (DA)~\cite{schoelkopf2012causal}. We will argue that this can explain performance differences reported by the NLP community across different data collection processes and tasks.
In particular, we make the following contributions:
\begin{enumerate}[itemsep=-4pt,topsep=1pt]
    \item We categorize a number of common NLP tasks according to the causal direction of the underlying data collection process~(\cref{sec:causal_categorization}).
    \item We review the ICM principle and its implications for common techniques of using unlabelled data such as SSL and DA
    in the context of causal and anticausal NLP tasks~(\cref{sec:implications_of_causal_anticausal_learning_for_NLP}).
    \item We empirically assay the validity of ICM for NLP data using minimum description length in a machine translation setting~(\cref{sec:mdl}).
    \item We verify experimentally and through a meta-study of over respectively 100 (SSL) and 30 (DA) published findings that the difference in SSL~(\cref{sec:ssl}) and domain adaptation (DA)~(\cref{sec:da}) performance on causal vs anticausal datasets reported in the literature
    is consistent with what is predicted by the ICM principle.
    \item We make suggestions on how to use findings in this paper for future work in NLP~(\cref{sec:how_to_use}).
\end{enumerate}

\section{Categorization of Common NLP Tasks into Causal and Anticausal Learning}
\label{sec:causal_categorization}

We start by categorizing common NLP tasks which use an input variable $X$ to predict a target or output variable $Y$ into causal learning ($X\rightarrow Y$), anticausal learning ($Y\rightarrow X$), and other tasks that do not have a clear underlying causal direction, or which typically rely on mixed (causal and anticausal)
types of data, as summarised in~\cref{tab:nlp_task_class}.

\newcommand{\ra}[1]{\renewcommand{\arraystretch}{#1}}
\begin{table}[t]
    \vspace{-0.5em}
    \centering
    \ra{1.2}
    \small
    \resizebox{\columnwidth}{!}{%
    \begin{tabular}{m{3cm}m{4.25cm}l}
    \toprule
    \textbf{Category} & \textbf{Example NLP Tasks} \\ \midrule
    \textbf{Causal learning} & Summarization, parsing, tagging, data-to-text generation, information extraction \\ \hline
    \textbf{Anticausal learning} & Author attribute classification, review sentiment classification \\ \hline
    \textbf{Other/mixed (depending on data collection)} & Machine translation, question answering, question generation, text style transfer, intent classification \\
    \bottomrule
    \end{tabular}
    }
    \caption{Classification of typical NLP tasks into causal (where the model takes the cause as input and predicts the effect), and anticausal (where the model takes the effect as input and predicts the cause) learning problems, as well as other tasks which do not have a clear causal interpretation of the data collection process, or where a mixture of both types of data is typically used.}
    \label{tab:nlp_task_class}
    \vspace{-0.5em}
\end{table}

Key to this categorization is determining whether the input $X$ corresponds to the cause or the effect in the data collection process.
As illustrated in~\cref{fig:prompt},
if the input $X$ and output $Y$ are generated at two different time steps, then the variable that is generated first is typically the cause, and the other that is subsequently generated
 is typically the effect, provided it is generated based on the previous one (rather than, say, on a common confounder that causes both variables).
If $X$ and $Y$ are generated jointly, then we need to distinguish based on the underlying generative process whether one of the two variables is causing the other variable.

\myparagraph{Learning Effect from Cause (Causal Learning)}
Causal ($X\rightarrow Y$) NLP tasks typically aim to predict a post-hoc generated human annotation (i.e., the target $Y$ is the effect) from a given input $X$ (the cause).
Examples include: summarization (\textit{article$\rightarrow$summary}) where the goal is to produce a summary $Y$ of a given input text $X$;
parsing and tagging (\textit{text$\rightarrow$linguists' annotated structure}) where the goal is to predict an annotated syntactic structure $Y$ of a given input sentence $X$; data-to-text generation (\textit{data$\rightarrow$description}) where the goal is to produce a textual description $Y$ of a set of structured input data $X$; 
and information extraction (\textit{text$\rightarrow$entities/relations/etc}) where the goal is to extract structured information
from a given text.

\myparagraph{Learning Cause from Effect (Anticausal Learning)}
Anticausal ($Y\rightarrow X$) NLP tasks typically aim to predict or infer some latent target property $Y$ such as an unobserved prompt from an observed input $X$ which takes the form of one of its effects.
Typical anticausal NLP learning problems include, for example, author attribute identification (\textit{author attribute$\rightarrow$text}) where the goal is to predict some unobserved attribute $Y$ of the writer of a given text snippet $X$;
and review sentiment classification (\textit{sentiment$\rightarrow$review text})  where the goal is to predict the latent sentiment $Y$ that caused an author to write a particular review $X$.

\myparagraph{Other/Mixed}
Some tasks can be categorized as either causal or anticausal, depending on how exactly the data is collected.
In~\cref{sec:intro}, we discussed the example of MT where different types of (causal and anticausal) data are typically mixed.
Another example is the task of intent classification: if 
the \textit{same} author reveals their intent before the writing (i.e., \textit{intent$\rightarrow$text}),  it can be viewed as an anticausal learning task; if, on the other hand, the data is annotated by \textit{other} people who are not the original author (i.e., \textit{text$\rightarrow$annotated intent}),  it can be viewed as a causal learning task.
A similar reasoning applies to question answering and generation tasks which respectively aim to provide an answer to a given question, or vice versa: if first a piece of informative text is selected and annotators are then asked to come up with a corresponding question (\textit{answer$\rightarrow$question}) as, e.g., in the SQuAD dataset~\cite{rajpurkar-etal-2016-squad}, then question answering is an anticausal and question generation a causal learning task; if, on the other hand, a question such as a search query is selected first and subsequently an answer is provided (\textit{question$\rightarrow$answer}) as, e.g., in the Natural Questions dataset~\cite{kwiatkowski2019natural}, then question answering is a causal and question generation an anticausal learning task. 
Often, multiple such datasets are combined without regard for their causal direction.

\section{Implications of ICM for Causal and Anticausal Learning Problems}
\label{sec:implications_of_causal_anticausal_learning_for_NLP}

\begin{figure*}[t]
    \vspace{-1em}
    \centering
    \includegraphics[width=\textwidth]{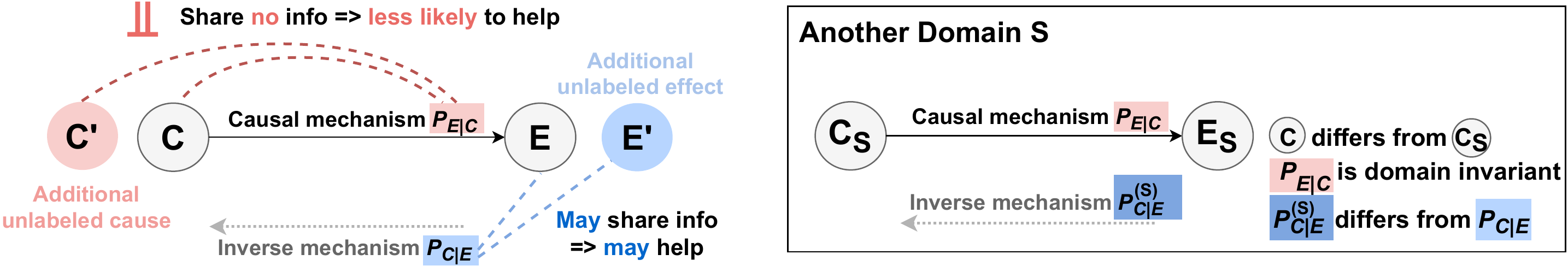}
    \vspace{-1.em}
    \caption{
    The ICM principle
    assumes that \textit{the generative process $P_C$ of the cause $C$ is independent of the causal mechanism $P_{E|C}$}: the two distributions share no information and each may be changed or manipulated without affecting the other. In the anticausal direction, on the other hand, the effect distribution \textit{$P_E$ is (in the generic case) not independent of the inverse mechanism $P_{C|E}$}: they may share information and change dependently.
    \textit{(Left)} SSL, which aims to improve an estimate of the target conditional $P_{Y|X}$ given additional unlabelled input data from $P_X$, should therefore not help for causal learning ($X\rightarrow Y$), 
    but may help in the anticausal direction ($Y\rightarrow X$).
    \textit{(Right)} DA,
    which aims to
    adapt a model of $P_{Y|X}$ from a source domain to a target domain (e.g., fine-tuning on a smaller dataset), should work better for causal learning settings where a change in $P_C$ is not expected to lead to a change in the mechanism $P_{E|C}$, whereas in the anticausal direction $P_E$ and $P_{C|E}$ may change in a \textit{dependent} manner.
    }
    \label{fig:icm}
    \vspace{-0.5em}
\end{figure*}

Whether we are in a causal or anticausal learning scenario has important implications for  semi-supervised learning (SSL)
and domain adaptation (DA)~\cite{schoelkopf2012causal,sgouritsa2015inference,zhang2013domain,zhang2015multi,gong2016domain,kugelgen2019semi,kugelgen2020semi}, which are techniques also commonly used in NLP.
These implications are derived from the principle of independent causal mechanisms (ICM)~\cite{schoelkopf2012causal,lemeire2006causal} which states that ``\textit{the causal generative process of a system's variables is composed of autonomous modules that do not inform or influence each other}''~\cite{peters2017elements}.

In the bivariate case, this amount to a type of independence assumption between 
the distribution~$P_C$ of the cause~$C$, and the causal process, or mechanism,%
~$P_{E|C}$ that generates the effect from the cause. 
For example, for a question answering task, the generative process~$P_C$ by which one person comes up with a question~$C$ is ``independent'' of the process $P_{E|C}$ by which another person produces an answer~$E$ for question~$C$.\footnote{The validity of this is meant in an approximate sense, and one can imagine settings where it is questionable. E.g., if the person asking the question has prior knowledge of the respondent (e.g., in a classroom setting), then she might adjust the question accordingly which would violate the assumption.}

Here, ``independent'' is not meant in the sense of \textit{statistical} independence of random variables, but rather as \textit{independence at the level of generative processes or distributions} in the sense that $P_C$ and $P_{E|C}$ \textit{do not share information} (the person asking the question and the one answering may not know each other) and \textit{can be manipulated independently of each other} (we can swap either of the two for another participant without the other one being influenced by this). 
Crucially, this type of independence is generally violated in the opposite, i.e., \textit{anticausal}, direction: $P_E$ and $P_{C|E}$ may share information and change dependently~\cite{DanJanMooZscSteZhaSch10,janzing2012information}.
This has two important implications for common learning tasks~\cite{schoelkopf2012causal} which are illustrated in~\cref{fig:icm}. 

\myparagraph{Implications of ICM for SSL}
First, if $P_C$ shares no information with $P_{E|C}$, SSL---where one has additional unlabelled input data from $P_X$ and aims to improve an estimate of the target conditional $P_{Y|X}$---should not work in the causal direction ($X\rightarrow Y$), but may work in the anticausal direction ($Y\rightarrow X$), as $P_E$ and $P_{C|E}$ may share information.
Causal NLP tasks should thus be less likely  to show improvements over a supervised baseline when using SSL than anticausal tasks.

\myparagraph{Implications of ICM for DA}
Second, according to the ICM principle, the causal mechanism $P_{E|C}$ should be invariant to changes in the cause distribution $P_C$, so domain---specifically, covariate shift~\cite{shimodaira2000improving,sugiyama2012machine}---adaptation, where $P_X$ changes but $P_{Y|X}$ is assumed to stay invariant, should work in the causal direction, but not necessarily in the anticausal direction.
Hence, 
DA should be easier for causal NLP tasks 
than for anticausal NLP tasks.

\section{Investigating the Validity of ICM for NLP Data Using MDL}
\label{sec:mdl}
Traditionally, the ICM principle is thought of in the context of \textit{physical} processes or mechanisms, rather than \textit{social} or \textit{linguistic} ones such as language.
Since ICM amounts to an independence assumption that---while well motivated in principle---may not always hold in practice,\footnote{E.g., due to confounding influences from unobserved variables, or mechanisms which have co-evolved to be dependent} we now assay its validity on NLP data.

Recall, that ICM postulates a type of independence between $P_C$ and $P_{E|C}$. One way to formalize this uses Kolmogorov complexity  $K(\cdot)$ as a measure of algorithmic information, which can be understood as the length of the shortest program that computes a particular algorithmic object such as a distribution or a function~\cite{
solomonoff1964formal,
kolmogorov1965three}.
ICM then reads~\cite{janzing2010causal}:\footnote{Here, $\overset{+}{=}$ and $\overset{+}{\leq}$ hold up a constant due to the choice of a Turing machine in the definition of algorithmic information.}
\vspace{-0.2cm}
\begin{align}
\label{eq:ICM_komogorov}
\begin{split}
K(P_{C,E})&\overset{+}{=} K(P_C) + K(P_{E|C}) \\
&\overset{+}{\leq} K(P_E)+K(P_{C|E})\,.
\end{split}
\end{align}
In other words, the shortest description of the joint distribution $P_{C, E}$ corresponds to describing $P_C$ and $P_{E|C}$ separately (i.e., they share no information), whereas there may be redundant (shared) information in the non-causal direction such that a  separate description of $P_E$ and $P_{C|E}$ will generally be longer than that of the joint distribution $P_{C, E}$.

\subsection{Estimation by MDL}\label{sec:estimate_by_mdl}

Since Kolmogorov complexity is not computable \cite{li2008introduction}, we adopt a commonly used proxy, the minimum description length (MDL) \cite{grunwald2007minimum}, to test the applicability of ICM for NLP data.
Given an input, such as a collection of observations $\{(c_i,e_i)\}_{i=1}^n\sim P_{C,E}$, MDL returns the shortest codelength (in bits) needed to compress the input, as well as the
parameters needed to decompress it.
We use MDL to approximate~\eqref{eq:ICM_komogorov} as follows:
\begin{align}
    &\mathrm{MDL}(\cb_{1:n},\eb_{1:n})= \mathrm{MDL}(\cb_{1:n}) + \mathrm{MDL}(\eb_{1:n}|\cb_{1:n}) 
    \nonumber
    \\
   & \quad\quad  \leq \mathrm{MDL}(\eb_{1:n}) + \mathrm{MDL} (\cb_{1:n} | \eb_{1:n}),
\label{eq:mdl_ineq}
\end{align}%
where MDL($\cdot|\cdot$) denotes a conditional compression where the second argument is treated as ``free parameters'' which do not count towards the compression length of the first argument.
Eq.~\eqref{eq:mdl_ineq} can thus be interpreted as a comparison between two ways of compressing the same data $(\cb_{1:n},\eb_{1:n})$: either we first compress $\cb_{1:n}$ and then compress $\eb_{1:n}$ conditional on $\cb_{1:n}$, or vice versa.
According to the ICM principle, 
the first way should tend to be more ``concise'' than the second.

\subsection{Calculating MDL Using Machine Translation as a Case Study}
\label{sec:mdl_seq2seq}

To  empirically assess the validity
of ICM for NLP data using MDL as a proxy, we turn to MT as a case study.
We choose MT because the input and output spaces of MT are relatively symmetric, as opposed to other NLP tasks such as text classification where the input space is sequences, but the output space is a small set of labels. 

There are only very few studies which calculate MDL on NLP data, so we extend the method of~\citet{voita2020information} to calculate MDL using online codes~\cite{rissanen1984universal} for deep learning tasks \cite{blier2018description}. Since the original calculation method for MDL by~\citet{voita2020information} was developed for classification, we extend it to sequence-to-sequence (Seq2Seq) generation.
Specifically, given a translation dataset $D = \{(\xb_1, \yb_1), \dots, (\xb_n, \yb_n) \}$ of $n$ pairs of sentences $\xb_i$ with translation $\yb_i$, denote the size of the vocabulary of the source language 
by $V_x$, and the size of the vocabulary of the target language
by $V_y$.
In order to assess whether~\eqref{eq:mdl_ineq} holds, we need to calculate four different terms: two marginal terms $\mathrm{MDL}(\xb_{1:n})$ and $\mathrm{MDL}(\yb_{1:n})$, and two conditional terms $\mathrm{MDL}(\yb_{1:n}|\xb_{1:n})$ and $\mathrm{MDL}(\xb_{1:n}|\yb_{1:n})$.

\myparagraph{Codelength of the Conditional Terms}
To calculate the codelength of the two conditional terms,
we extend the method of \citet{voita2020information} from classification to Seq2Seq generation. Following the setting of \citet{voita2020information}, we break the dataset $D$ into 10 disjoint subsets with increasing sizes and denote the end index of each subset as~$t_i$.\footnote{The sizes of the 10 subsets are 0.1, 0.2, 0.4, 0.8, 1.6, 3.2, 6.25, 12.5, 25, and 50 percent of the dataset size, respectively. E.g., $t_1=0.1\% n, t_2 = (0.1\%+0.2\%) n, \dots$.
}
We then estimate  $\mathrm{MDL}(\yb_{1:n} |\xb_{1:n})$ as
\begin{align}
    & \widehat{\mathrm{MDL}}(\yb_{1:n} |\xb_{1:n}) = {\textstyle \sum_{i=1}^{t_1}} \mathrm{length}(\yb_i) \cdot
\log_2 V_y
\nonumber\\ 
& - {\textstyle\sum_{i=1}^{n-1}} \log_2 p_{\theta_i} (\yb_{1+t_i:t_{i+1}} | \xb_{1+t_i:t_{i+1}})
~, 
\label{eq:mdl_yx}
\end{align}
where $\mathrm{length}(\yb_i)$ refers to the number of tokens in the sequence $\yb_i$, $\theta_i$ are the parameters of a translation model $h_i$ trained on the first $t_i$ data points, and $\bm{\mathrm{seq}}_{\mathrm{idx}_1: \mathrm{idx}_2}$ refers to the set of sequences from the $\mathrm{idx}_1$-th to the $\mathrm{idx}_2$-th sample in the dataset $D$, where $\bm{\mathrm{seq}} \in \{\xb, \yb\}$ and $\mathrm{idx}_i \in \{1, \dots, n\}$.
Similarly, when calculating $\mathrm{MDL}(\xb_{1:n}|\yb_{1:n})$, we simply swap the roles of $\xb$ and $\yb$.

\begin{table}[t]
\ra{1.2}
    \centering
    \small
    \begin{tabular}{llll}
    \toprule
    \textbf{Dataset} & \textbf{Size} 
    & \textbf{Note}
    \\ \midrule
    En$\rightarrow$Es & 81K & Original English, Translated Spanish \\
    Es$\rightarrow$En & 81K & Original Spanish, Translated English \\
    En$\rightarrow$Fr & 16K & Original English, Translated French \\
    Fr$\rightarrow$En & 16K & Original French, Translated English \\
    Es$\rightarrow$Fr & 15K & Original Spanish, Translated French \\
    Fr$\rightarrow$Es & 15K & Original French, Translated Spanish \\
    \bottomrule
    \end{tabular}
    \vspace{-0.5em}
    \caption{Details of the CausalMT corpus.
    }
    \label{tab:causalmt}
    \vspace{-1em}
\end{table}

\begin{table*}[th!]
    \vspace{-1em}
    \centering
    \small
    \ra{1.1}
    \resizebox{\textwidth}{!}{
    \begin{tabular}{>{\centering\arraybackslash}m{2cm}cccc>{\centering\arraybackslash}m{6.5cm}}
    \toprule
    \textbf{Data (X$\rightarrow$Y)} & \textbf{MDL(X)} & \textbf{MDL(Y)} & \textbf{MDL(Y|X)} & \textbf{MDL(X|Y)} & \textbf{MDL(X)+MDL(Y|X)} 
    \textbf{vs. MDL(Y)+MDL(X|Y)} \\ \midrule
    En$\rightarrow$Es & 46.54 & 105.99 & 2033.95 & 2320.93 & 2080.49 $<$ 2426.92 \\
    Es$\rightarrow$En & 113.42 & 55.79 & 3289.99 & 3534.09 & 3403.41 $<$ 3589.88 \\
    En$\rightarrow$Fr & 20.54 & 53.83 & 503.78 & 535.88 & 524.32 $<$ 589.71 \\
    Fr$\rightarrow$En & 53.83 & 21.6 & 705.28 & 681.12 & 759.11 $>$ 702.72 \\
    Es$\rightarrow$Fr & 58.26 & 55.66 & 701.04 & 755.5 & 759.30 $<$ 811.16 \\
    Fr$\rightarrow$Es & 56.14 & 54.34 & 665.26 & 706.53 & 721.40 $<$ 760.87 \\
    \bottomrule
    \end{tabular}
    }
    \vspace{-0.5em}
    \caption{Codelength (in kbits) of $\mathrm{MDL}(X)$, $\mathrm{MDL}(Y)$, $\mathrm{MDL}(Y|X)$, and $\mathrm{MDL}(X|Y)$ on six CausalMT datasets.}
    \label{tab:mdl_res}
    \vspace{-1em}
\end{table*}

\myparagraph{Codelength of the Marginal Terms}
When calculating the two marginal terms, $\mathrm{MDL}(\xb_{1:n})$ and $\mathrm{MDL}(\yb_{1:n})$, we make two changes from the above calculation of conditional terms: first, we replace the \textit{translation models} $h_i$ with \textit{language models}; second, we remove the conditional distribution. That is, we calculate $\mathrm{MDL}(\xb_{1:n})$ as
\begin{equation}
\begin{split}
\widehat{\mathrm{MDL}}(\xb_{1:n}) & = 
{\textstyle\sum_{i=1}^{t_1}} \mathrm{length}(\xb_i) \cdot \log_2 V_x
\\ 
& - {\textstyle\sum_{i=1}^{n-1}} \log_2 p_{\theta_i} (\xb_{1+t_i:t_{i+1}} )
~,
\end{split}
\end{equation}
where $\theta_i$ are the parameters of a language model~$h_i$ trained on the first $t_i$ data points. We apply the same method to calculate $\mathrm{MDL}(\yb_{1:n})$.

For the language model, we use GPT2 \cite{radford2019language_gpt2}, and for the translation model, we use the Marian neural machine translation model \cite{mariannmt} trained on the OPUS Corpus \cite{tiedemann-nygaard-2004-opus}. 
For fair comparison, all models adopt the transformer architecture \cite{vaswani2017attention}, and 
have roughly the same number of parameters.
See Appendix~\ref{appd:mdl} for more experimental details.

\subsection{CausalMT Corpus}
\label{sec:mdl_mt}

For our MDL experiment,
we need datasets for which the causal direction of data collection is known, i.e., for which we have ground-truth annotation of which text is the original and which is a translation, instead of a mixture of both.
Since existing MT corpora do not have this property as discussed in~\cref{sec:intro}, we curate our own corpus, which we call the CausalMT corpus.

Specifically, we consider the existing MT dataset WMT'19,\footnote{\href{http://www.statmt.org/wmt19/parallel-corpus-filtering.html}{Link to WMT'19}.} and identify some subsets that have a clear notion of causality. The subsets we use are the EuroParl~\cite{koehn2005europarl} and Global Voices translation corpora.\footnote{\href{http://casmacat.eu/corpus/global-voices-tar-balls/training.tgz}{Link to Global Voices}.}
For EuroParl, each text has meta information such as the speaker's language; for Global Voices, each text has meta information about whether it is translated or not. We regard  text that is in the same language as the speaker's native language in EuroParl (and non-translated text in Global Voices) as the original (i.e., the cause). We then retrieve a corresponding effect by using the cause text to match the parallel pairs in the processed dataset.
In this way, we compile six translation datasets with clear causal direction as summarized in~\cref{tab:causalmt}. For each dataset, we use 1K samples each as test and validation sets, and use the rest for training.

\subsection{Results}
The results of our MDL experiment on the six CausalMT datasets are summarised in~\cref{tab:mdl_res}. 
If ICM holds, we expect the sum of codelengths to be smaller for the causal direction than for the anticausal one, see~\eqref{eq:mdl_ineq}.
As can be seen from the last column, this is the case for five out of the six datasets.
For example, on one of the  largest datasets (En$\rightarrow$Es), the MDL difference is 346 kbits.\footnote{
As far as we know, determining statistical significance in the investigated setting remains an open problem. While, in theory, one may use information entropy to estimate it, in practice, this may be inaccurate since (i) MDL is only a proxy for algorithmic information; and (ii) ICM may not hold exactly, but only approximately. We evaluate on six different datasets, so that the overall results can show a general trend.
}

Comparing the dataset sizes in~\cref{tab:causalmt} and results in~\cref{tab:mdl_res}, we observe that the absolute MDL values are roughly proportional to  dataset size, but other factors such as language and task complexity also play a role.
This is inherent to the nature of MDL being the sum of codelengths of the model and of the data given the model. 
Since we use equally-sized datasets for each language pair in the CausalMT corpus (i.e., in both the $X\rightarrow Y$ and $Y\rightarrow X$ directions, see~\cref{tab:causalmt}), numbers for the same language pair in~\cref{tab:mdl_res}, including the most important column ``MDL(X)+MDL(Y|X) vs.\ MDL(Y)+MDL(X|Y)'', 
form a valid comparison. 
That is, En\&Es experiments are comparable within themselves, so are the other language pairs.

For some of the smaller differences in the last column in~\cref{tab:mdl_res}, and, in particular the reversed inequality in row 4, a potential explanation may be the relatively small dataset size, as well as the fact that text data may be confounded (e.g., through shared grammar and semantics).

\section{SSL for Causal vs. Anticausal Models}\label{sec:ssl}
In semi-supervised learning (SSL), we are given a typically-small set of $k$ labeled observations $D_L = \{(\bm{x}_1, \bm{y}_1), \dots, (\bm{x}_k, \bm{y}_k) \}$, and a typically-large set of $m$ unlabeled observations of the input $D_U = \{\bm{x}_1^{(u)}, \dots, \bm{x}_m^{(u)}\}$.
SSL then aims to use the additional information about the input distribution
$P_X$
from the unlabeled dataset $D_U$ to improve a model of $P_{Y|X}$ learned on the labeled dataset $D_L$.

As explained in~\cref{sec:implications_of_causal_anticausal_learning_for_NLP}, SSL should only work for anticausal (or confounded) learning tasks, according to the ICM principle.
\citet{schoelkopf2012causal} have observed this trend on a number of classification and regression tasks on small-scale numerical inputs, such as predicting Boston housing prices from quantifiable neighborhood features (causal learning), or breast cancer from lab statistics (anticausal learning). 
However, there exist no studies investigating the implications of ICM for SSL on NLP data, which is of a more complex nature due to the high dimensionality of the input and output spaces, as well as potentially large confounding. In the following, we use a sequence-to-sequence decipherment experiment~(\cref{sec:decipherment}) and a meta-study of existing literature~(\cref{sec:meta_ssl}) to showcase that the same phenomenon also occurs in 
NLP.

\subsection{Decipherment Experiment}
\label{sec:decipherment}
To have control over causal direction of the data collection process, we use a synthetic decipherment dataset to test the difference in  SSL improvement between causal and anticausal learning tasks.

\myparagraph{Dataset}
We create a synthetic dataset of encrypted sequences. Specifically, we (i) adopt a monolingual English corpus (for which we use the English corpus of the En$\rightarrow$Es in the CausalMT dataset, for convenience), (ii) apply the ROT13 encryption algorithm~\cite{schneier1996applied} to obtain the encrypted corpus, and  then (iii) apply noise on the corpus that is chosen to be the effect corpus.

In the encryption step (ii), for each English sentence $\bm{x}$, its encryption $\mathrm{ROT13}(\bm{x})$ replaces each letter with the 13th letter after it in the alphabet, e.g., ``A''$\rightarrow$``N,'' ``B''$\rightarrow$``O.''
Note that we choose ROT13 due to its invertibility, since $\mathrm{ROT13}(\mathrm{ROT13}(\bm{x}))=\bm{x}$.
Therefore, without any noises, the corpus of English and the corpus of encrypted sequences by ROT13 are symmetric.

In the noising step (iii), we apply noise either to the English text or to the ciphertext, thus creating two datasets Cipher$\rightarrow$En, and En$\rightarrow$Cipher, respectively.
When applying noise to a sequence, we use the implementation of the Fairseq library.\footnote{
\href{https://github.com/pytorch/fairseq/blob/master/fairseq/data/denoising\_dataset.py}{Link to the Fairseq implementation}.
}
Namely, we mask some random words in the sequence (word masking), permute a part of the sequence (permuted noise), randomly shift the endings of the sequence to the beginning (rolling noise), and insert some random characters or masks to the sequence (insertion noise). We set the probability  of all noises to $p=5\%$. 

\begin{table}[t]
\vspace{-1em}
    \centering
    \ra{1.1}
    \small
    \resizebox{\columnwidth}{!}{%
    \begin{tabular}{lllll}
    \toprule
    \textbf{Causal Data} & \textbf{Learning Task} & \textbf{Sup. BLEU} & \textbf{$\Delta$SSL (BLEU)} \\ \midrule
    \multirow{2}{*}{En$\rightarrow$Cipher} & Causal
    & 19.20 & +1.84 \\
    & Anticausal & 7.75 & +38.02 \\ \hline
    \multirow{2}{*}{Cipher$\rightarrow$En} & Causal
    & 17.08 & +4.05 \\
    & Anticausal & 7.97 & +38.01 \\ 
    \bottomrule
    \end{tabular}
    }
    \vspace{-0.5em}
    \caption{SSL improvements ($\Delta$SSL) in BLEU score across causal vs.\ anticausal learning tasks on the synthetic decipherment datasets.
    }
    \label{tab:cipher}
    \vspace{-1em}
\end{table}

\myparagraph{Results}
For each of the two datasets En$\rightarrow$Cipher and Cipher$\rightarrow$En, we perform SSL in the causal and anticausal direction by either treating the input $X$ as the cause and the target $Y$ as the effect, or vice versa.
Specifically, we use a standard Transformer architecture for the supervised model, and for SSL, we multitask the translation task with an additional denoising autoencoder~\cite{vincent2008extracting} using the Fairseq Python package.
The results are shown
in~\cref{tab:cipher}.
It can be seen that in both cases, anticausal models show a substantially larger SSL improvement  than causal models.

We also note that there is a substantial gap in the supervised performance between causal and anticausal learning tasks on the same underlying data. 
This is also expected as 
causal learning is typically easier than anticausal learning since it corresponds to learning the ``natural'' forward function, or causal mechanism, while anticausal learning corresponds to learning the less natural, non-causal inverse mechanism.

\subsection{SSL Improvements in Existing Work}\label{sec:meta_ssl}

After verifying the different behaviour in SSL improvement predicted by the ICM principle on the decipherment experiment, we conduct an extensive meta-study to survey whether this trend is also reflected in published NLP findings.
To this end, we consider a diverse set of tasks, and SSL methods. The tasks covered in our meta-study include machine translation, summarization, parsing, tagging, information extraction, review sentiment classification, text category classification, word sense disambiguation, and chunking. The SSL methods include self-training, co-training~\cite{blum1998combining}, tri-training~\cite{zhou2005tri}, transductive support vector
machines~\cite{joachims1999transductive}, expectation maximization~\cite{nigam2006semi}, multitasking with language modeling~\cite{dai2015semi}, multitasking with sentence reordering (as used in~\citet{zhang-zong-2016-exploiting}), and cross-view training~\cite{clark2018semi}. Further details on our meta study are explained in Appendix~\ref{appd:meta_ssl}.

\begin{table}[t]
\vspace{-1em}
    \centering
    \ra{1.1}
    \small
    \resizebox{\columnwidth}{!}{%
    \begin{tabular}{lclc}
    \toprule
    \textbf{Task Type} & \textbf{Mean} \textbf{$\Delta$SSL ($\pm$std)} & \textbf{According to ICM} \\ \midrule
    Causal & +0.04 ($\pm$4.23) & Smaller or none \\
    Anticausal & +1.70 ($\pm$2.05) & Larger \\
    \bottomrule
    \end{tabular}
    }
    \vspace{-0.5em}
    \caption{Meta-study of SSL improvement ($\Delta$SSL) across 55  causal and 50 anticausal NLP tasks. 
    }
    \label{tab:meta_ssl}
    \vspace{-1em}
\end{table}

We covered 55 instances of causal learning and 50 instances of anticausal learning. A summary of the trends of causal SSL and anticausal SSL are listed in \cref{tab:meta_ssl}. Echoing with the implications of ICM stated in~\cref{sec:implications_of_causal_anticausal_learning_for_NLP}, for causal learning tasks, the average improvement by SSL is only very small, 0.04\%. In contrast, the anticausal SSL improvement is larger, 1.70\% on average. We use Welch's t-test~\cite{welch1947generalization} to assess whether the difference in mean between the two distributions of SSL improvment (with unequal variance) is significant 
and obtain a p-value of 0.011.

\section{DA for Causal vs. Anticausal Models}\label{sec:da}

We also consider a supervised domain adaptation (DA) setting in which the goal is to adapt a model trained on a  large labeled data set 
from a source domain, to a potentially different target domain from which we only have a
a small labeled data set.
As explained in~\cref{sec:implications_of_causal_anticausal_learning_for_NLP}, DA 
should only work well for causal learning, but not necessarily for anticausal learning, according to the ICM principle.

Similar to the meta-study on SSL, we also review existing NLP literature on DA. We focus on DA improvement, i.e., the performance gain of using DA over an unadapted baseline that only learns from the source data and is tested on the target domain. 
Since the number of studies on DA that we can find is smaller than for SSL, we cover 22 instances of DA on causal tasks, and 11 instances of DA on anticausal tasks.

\begin{table}[t]
\small\ra{1.1}
    \centering
    \begin{tabular}{lclc}
    \toprule
    \textbf{Task Type} & \textbf{Mean} $\Delta$\textbf{DA} ($\pm$std) & \textbf{According to ICM} \\ \midrule
    Causal & 5.18 ($\pm$6.57) & Larger \\
    Anticausal & 1.26 ($\pm$1.79) & Smaller \\
    \bottomrule
    \end{tabular}
    \vspace{-0.5em}
    \caption{Meta-study of DA improvement ($\Delta$DA) across 22 causal and 11 anticausal NLP tasks.
    }
    \label{tab:da}
    \vspace{-1em}
\end{table}

The results are summarised in~\cref{tab:da}. We find that the observations again echo with our expectations (according to ICM) that DA should work better for causal, than for anticausal learning tasks.
Again, we use Welch's t-test~\cite{welch1947generalization} to verify that the DA improvements of causal learning and anticausal learning are statistically different, and obtain a p-value of 0.023.

\section{How to Use the Findings in this Study}\label{sec:how_to_use}
\myparagraph{{Data Collection Practice in NLP}}
{Due to the different implications of causal and anticausal learning tasks, \textit{we strongly suggest annotating the causal direction when collecting new NLP data.}
One way to do this is to only collect data from one causal direction and to mention this in the meta information.
For example, summarization data collected from the TL;DR of scientific papers \text{SciTldr}~\cite{cachola-etal-2020-tldr} should be \textit{causal}, as the TL;DR summaries on OpenReview (some from authors when submitting the paper, others derived from the beginning of peer reviews) were  likely composed after the original papers or reviews were written.
Alternatively, one may allow mixed corpora, but label the causal direction for each $(\bm{x}, \bm{y})$ pair, e.g., which is the original vs.\ translated text in a translation pair.
Since more data often leads to better model performance, it is common to mix data from both causal directions, e.g., training on both En$\rightarrow$Es and Es$\rightarrow$En data.
Annotating the causal direction for each pair allows future users of the dataset to potentially 
handle the causal and anticausal parts of the data differently.
}

\myparagraph{Causality-Aware Modeling}
{
When building NLP models, the causal direction provides additional information that can
potentially be built into the model.
In the MT case, since
causal and anticausal learning can lead to different
performance~\citep{ni2021original}, 
one way to take advantage of the known causal direction is to add a prefix such as ``[Modeling-Effect-to-Cause]'' to the original input, so that the model can learn from causally-annotated input-output pairs. For example,~\citet{riley-etal-2020-translationese} use labels of the causal direction
to elicit different behavior at inference time.
Another option is to carefully design a combination of different modeling techniques, such as limiting self-training (a method for SSL) only to the anticausal direction and allowing back-translation in both directions, as preliminarily explored by~\citet{shen-etal-2021-source}.
}

\myparagraph{Causal Discovery}
{
Suppose that we are given measurements of two types of NLP data $X$ and $Y$ (e.g., text, parse tree, intent type) whose collection process is unknown, i.e., which is the cause and which the effect. 
One key finding of our study is that there is typically a causal footprint of the data collection process which manifests itself, e.g., when computing the description length in different directions~(\cref{sec:mdl}) or when performing SSL~(\cref{sec:ssl}) or DA~(\cref{sec:da}). Based on which direction has the shorter MDL, or allows better SSL or DA, we can thus infer one causal direction over the other.
}

\myparagraph{Prediction of SSL and DA Effectiveness}
Being able to predict the effectiveness of SSL or DA for a given NLP task can be very useful, e.g., to set the weights in an ensemble of different models~\cite{sogaard2013semi}.
While predicting SSL performance has previously been studied from a non-causal perspective~\cite{nigam2000analyzing, asch2016predicting}, our findings suggest that a simple qualitative description of the data collection process in terms of its causal direction (as summarised for the most common NLP tasks in~\cref{tab:nlp_task_class}) can also be surprisingly effective to evaluate whether SSL or DA should be expected to work well.

\section{{Limitations and Future Work}}
{
We note that ICM---when taken strictly---is an idealized
assumption that may be violated and thus may not hold exactly for a given
real-world data set, e.g., due to 
confounding, i.e., when both variables are influenced by a third, unobserved variable.
In this case, one may observe less of a difference between causal and anticausal learning tasks.
}

{We also note that, while we have made an effort to classify different NLP tasks as \textit{typically} causal or anticausal, our categorization should not be applied blindly without regard for the specific generative process at hand: deviations are possible as explained in the Mixed/Other category.}

Another limitation is that the SSL and DA settings considered in this paper are only a subset of the various settings that exist in NLP. Our study does not cover, for example, SSL that uses additional output data (e.g.,~\citet{jean2015montreal,gulcehre2015using,sennrich-zhang-2019-revisiting}), or unsupervised DA (as reviewed by~\citet{ramponi-plank-2020-neural}).
In addition, in our meta-study of published  SSL and DA findings, the improvements of causal vs.\ anticausal learning might be amplified by the scale of research efforts on different tasks and potentially suffer from selection bias.

{Finally, we remark that, in the present work, we have focused on bivariate prediction tasks with an input $X$ and output $Y$. 
Future work may also apply ICM-based reasoning to more complex NLP settings, for example, by (i) incorporating additional (sequential/temporal) structure of the data (e.g., for MT or language modeling) or (ii) considering settings in which the input $X$ consists of both cause $X_\textsc{cau}$ and effect $X_\textsc{eff}$ features of the target $Y$~\citep{kugelgen2019semi,kugelgen2020semi}.}

\section{Related Work}

\noindent\textbf{NLP and Causality}
Existing work on NLP and causality mainly focuses on the extracting text features  for causal inference. Researchers first propose a causal graph based on domain knowledge, and then use text features to represent some elements in the causal graph, e.g., the cause~\cite{egami2018make,jin2021causal}, effect~\cite{fong2016discovery}, and confounders~\cite{roberts2020adjusting,veitch2020adapting,keith2020text}. Another line of work mines causal relations among events from textual expressions, and uses them to perform relation extraction~\cite{do-etal-2011-minimally,mirza-tonelli-2014-analysis,dunietz-etal-2017-corpus,hosseini2021predictting}, question answering~\cite{oh2016semi}, or commonsense reasoning~\cite{sap2019atomic,bosselut-etal-2019-comet}.
For a recent survey, we refer to~\citet{feder2021causal}.

\myparagraph{Usage of MDL in NLP}
Although MDL has been used for causal discovery for low-dimensional data~\cite{budhathoki2017mdl,mian2021discovering,marx2021formally}, only very few studies adopt MDL on high-dimensional NLP data. Most existing uses of MDL on NLP are for probing and interpretability: e.g.,
\citet{voita2020information} use it for probing of a small Bayesian model and network pruning, based on the method proposed by~\citet{blier2018description} to calculate MDL for deep learning.
We are not aware of existing work using MDL for causal discovery, or to verify causal concepts such as ICM in the context of NLP.

\myparagraph{Existing Discussions on SSL and DA in NLP}
SSL and DA has long been used in NLP, as reviewed by~\citet{sogaard2013semi} and~\citet{ramponi-plank-2020-neural}. However, there have been a number of studies that report negative results for SSL~\cite{clark-etal-2003-bootstrapping,steedman-etal-2003-bootstrapping,reichart-rappoport-2007-self,abney2007semisupervised,spreyer-kuhn-2009-data,sogaard-rishoj-2010-semi} and DA~\cite{plank2014importance}. 
Our works constitutes the first explanation of  the ineffectiveness of SSL and DA on certain NLP tasks from the perspective of causal and anticausal learning.

\section{Conclusion}
This work presents the first effort to use causal concepts such as the ICM principle and the distinction between causal and anticausal learning to shed light on some commonly observed trends in NLP.
Specifically, we provide an  explanation of observed differences in SSL~(\cref{tab:cipher,tab:meta_ssl}) and DA~(\cref{tab:da}) performance on a number of NLP tasks: DA tends to work better for causal learning tasks, whereas SSL typically only works for anticausal learning tasks, as predicted by the ICM principle.
These insights, together with our categorization of common NLP tasks~(\cref{tab:nlp_task_class}) into causal and anticausal learning, may prove useful for future NLP efforts.
Moreover, we empirically confirm using MDL that the description of data is typically shorter in the causal than in the anticausal direction~(\cref{tab:mdl_res}), suggesting that a causal footprint can also be observed for text data. This has interesting potential implications for discovering causal relations between different types of NLP data.

\section*{Acknowledgements}
We thank Simon Buchholz for helpful discussions, Nasim Rahaman, Shehzaad Dhuliawala, Yifan Hou, Tiago Pimentel, and the anonymous reviewers for feedback on the manuscript, and Di Jin for helping with computational resources. 
This work was supported by the German Federal Ministry of Education and Research (BMBF): Tübingen AI Center, FKZ: 01IS18039B, and by the Machine Learning Cluster of Excellence, EXC number 2064/1 – Project number 390727645.
\section*{Ethical Considerations}

\myparagraph{Use of Data} 
This paper uses two types of data, a subset of an existing machine translation dataset, and synthetic decipherment data. As far as we know, there are no sensitive issues such as privacy regarding the data usage.

\myparagraph{Potential Stakeholders}
This research focuses on meta properties of two commonly applied methodologies, SSL and DA in NLP. Although this research is not directly connected to specific applications in society, the usage of this study can benefit future research in SSL and DA.

\bibliography{anthology,custom}
\bibliographystyle{acl_natbib}

\newpage
\clearpage
\appendix

\section{Meta Study Settings of SSL and DA}\label{appd:meta_ssl}
For the meta study of SSL, we covered but are not limited to all relevant papers cited by the review on NLP SSL by \citet{sogaard2013semi}. We went through the leaderboard of many NLP tasks and covered the SSL papers listed on the leaderboards.
The papers covered by our meta study are available on our GitHub.

For supervised DA, we searched papers with the keyword domain adaptation and task names from a wide range of tasks that use supervised DA.

Note that for fair comparison, we do not consider papers without a comparable supervised baseline corresponding to the SSL, or a comparable unadapted baseline corresponding to the DA.
We do not consider MT DA which tackles the out-of-vocabulary (OOV) problem because $P(E|C)$ may be different for OOV \cite{habash2008four,daume2011domain}.

\section{Experimental Details of Minimum Description Length}\label{appd:mdl}
We calculate the MDL(X) and MDL(Y) by a language model, and obtain MDL(X|Y) and MDL(Y|X) using translation models. For language model, we use the autoregressive GPT2 \cite{radford2019language_gpt2}, and for the translation model, we the Marian Neural Machine Translation model \cite{mariannmt} trained on the OPUS Corpus \cite{tiedemann-nygaard-2004-opus}. Both these models use the layers from the transformer model \cite{vaswani2017attention}. The autoregressive language model consists only of decoder layers, whereas the translation model used six encoder and six decoder layers. Both of these models have roughly the same number of parameters. We used the huggingface implementation \cite{wolf-etal-2020-transformers} of these models for their respective set of languages.

\end{document}